\documentclass[11pt,a4paper]{article}
\usepackage[hyperref]{acl2019}
\usepackage{times}
\usepackage{latexsym}

\usepackage{url}

\aclfinalcopy 

\title{When is ACL's Deadline? A Scientific Conversational Agent}

\author{
Mohsen Mesgar,
Paul Youssef,
Lin Li, 
Dominik Bierwirth, \\
\textbf{Yihao Li, 
Christian M. Meyer,} and 
\textbf{Iryna  Gurevych}\\
Ubiquitous Knowledge Processing Lab (UKP) \\
Department of Computer Science  \\
Technische Universit{\" a}t Darmstadt\\
\url{https://www.ukp.tu-darmstadt.de}
}

\date{}
\usepackage{color}
\usepackage{tikz}
\usetikzlibrary{shapes}
\usepackage{booktabs}
\usepackage[linguistics]{forest} 

\newcommand{\athena}{\resizebox{4mm}{4mm}{\includegraphics[]{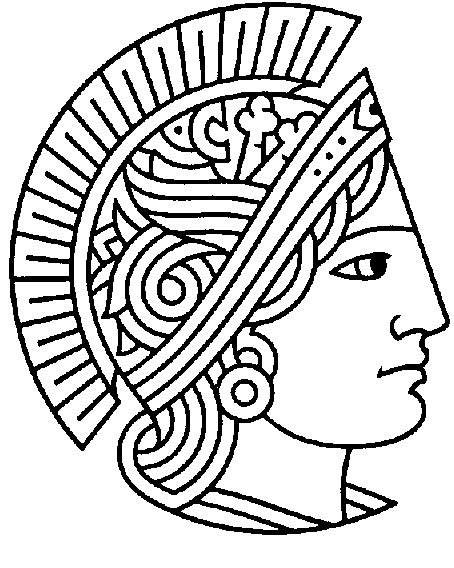}}}

\begin{document}
\maketitle
\begin{abstract}
Our conversational agent UKP-ATHENA assists NLP researchers in finding and exploring scientific literature, identifying relevant authors, planning or post-processing conference visits, and preparing paper submissions using a unified interface based on natural language inputs and responses. 
UKP-ATHENA enables new access paths to our swiftly evolving research area with its massive amounts of scientific information and high turnaround times. 
UKP-ATHENA's responses connect information from multiple heterogeneous sources which researchers currently have to explore manually one after another. 
Unlike a search engine, UKP-ATHENA maintains the context of a conversation to allow for efficient information access on papers, researchers, and conferences. 
Our architecture consists of multiple components with reference implementations that can be easily extended by new skills and domains. 
Our user-based evaluation shows that UKP-ATHENA already responds 45\% of different formulations of defined intents with 37\% coverage rate. 
\end{abstract}

%
%
%
%

\section{Introduction}
Researchers need to be \mbox{up-to-date} about the latest status of their research areas to deliver novel contributions. 
However, the amount of such information is exploding as research areas are rapidly growing in various aspects such as the number of published papers, authors,  conferences, conference participants etc. 
%
%

%
Several solutions have been proposed to obtain new insights from such heterogeneous information.
%
%
%
%
For example, GrapAL\footnote{\url{https://allenai.github.io/grapal-website/}} \cite{betts2019grapal} is a web-based tool for exploring scientific literature enabling, e.g., finding experts on a given topic for peer reviewing. 
%
Google and semantic Scholar are two web-based tools that provide information about researchers (e.g.\ their h-index) and papers (e.g.\ the number of their citations). 
Some tools are specifically designed for the NLP research area:  
%
the ACL Anthology is one of the primary knowledge bases that collects papers published in the NLP conferences and journals. 
CL Scholar \cite{singh2018cl} develops a knowledge graph from the ACL Anthology. 
\newcite{wan2019aminer} propose solutions for expert finding, trend analysis, reviewer recommendation and alike. 

Despite the valuable outputs of these solutions, their benefits remain restricted as they are not working together under a unique interaction environment.  
Each of these tools has its own user-interface, which is not consistent with those of other tools.  
None of them interacts with users via human (natural) language. 
Moreover, the insights provided by these disjoint solutions are independent of the history of interactions with users. 

Recent advances in conversational agents have shown to simplify the interactions between human users and computers in various tasks such as chitchats \cite{serban2017deep}, recommending a restaurant \cite{wen-etal-2017-network}, and booking a table \cite{bordes2016learning}. 
%
%
%
While such agents become available to consumers at a large scale, the NLP and ML research community who largely contributes to the agents' development does not yet use this technique for boosting the scientific process. 
%
\begin{figure*}[!ht]
    \centering
    \resizebox{\textwidth}{!}
    {
    \input{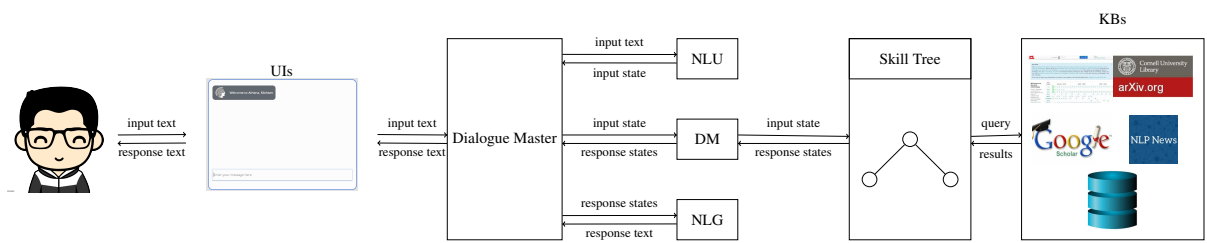}
    }
    \caption{A general view of the components used in \athena.}
    \label{fig:athena_architecture}
\end{figure*}
In this paper, we present UKP-ATHENA (henceforth~\athena) as a scientific conversational agent, which provides easy access to massive scientific information.  
\athena\ responds to questions about various aspects of the NLP research area by retrieving answers from multiple scientific knowledge bases and services.  
%
To the best of our knowledge, \athena\ is the first open-source conversational agent developed for helping researchers in finding and exploring scientific literature, identifying relevant authors, planning conference visits and preparing paper submissions. 
%

%
%
We perform a human evaluation to measure the quality of dialogues between \athena\ and researchers.   
Participant researchers in our study find that \athena\ is  beneficial for their research because it already fulfils many of their essential needs.
%
%
%

\section{UKP-ATHENA}
The general architecture of \athena\ is shown in Figure~\ref{fig:athena_architecture}. 
%
%
Its modules are grouped into the user interfaces (UIs), the main dialogue components natural language understanding (NLU), dialogue management (DM), and natural language generation (NLG) directed by a master, to query the requested information. 
The DM is backed by a tree of multiple skills returning information from external Knowledge Bases (KBs) or services. 
%
%
\subsection{User Interfaces (UIs)}
User interfaces (UIs) let users interact with machines easily.  
We implement a web-form, a command-based, and a web-service UI for \athena. 
The web-form UI is appealing to interact with for non-technical human users, and the command-based UI fits developers and technical users. 
The web-service UI enables \athena\ to be used as a virtual member in chat-rooms.
%
%
%

\subsection{Dialogue Master}
Dialogue master is responsible to communicate with the three main dialogue components for which it uses our primary data structure ``state''. 
A state encodes a dialogue state which includes salient information presented in any utterance: \emph{domain}, \emph{intent}, and \emph{slots}. 
If information in a state is extracted from an utterance said by a user, we refer to it as an \emph{input state}.
If it is provided by \athena\ for generating a response, we refer to it as a \emph{response state}. 

\begin{figure*}[!ht]
    \resizebox{\textwidth}{!}{
\begin{forest}
  [Master
    [General
     [Context]
     [Exit
        [Survey]
     ]
     [Fallback]
     [Feedback]
     [Greeting]
     [Identity]
     [Menu]
    ]
    [Task
    [Paper
        [Meta-data]
        [Discourse]
    ]
    [Conference
        [Dates
            [Deadlines]
        ]
        [Events
            [Keynotes]
            [Social\_events]
            [Tutorials]
        ]
        [Program]
    ]
    [People]
    [NLPnews]
    ]
  ]
\end{forest}
}
    \caption{The tree hierarchical relationships among domains implemented for the current version of \athena.}
    \label{fig:skills}
\end{figure*}
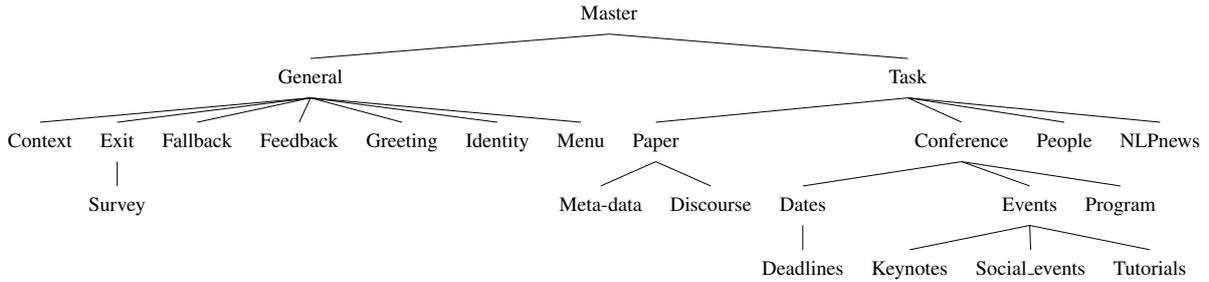

A domain indicates the topic of an utterance, e.g., \emph{conference}, \emph{paper}, and \emph{people}. 
An intent refers to the intention of the speaker of saying an utterance, e.g., \emph{give-deadlines} is an intent in the \emph{conference} domain. 
Slots are lists of nominal entities that are required to fulfill an intent in a domain, e.g., \{CONF-NAME\} is a slot for intent \emph{give-deadlines} in domain \emph{conference}. 
%
The implementation of \athena\ released with this paper has $46$ intents and $11$ slots for domains shown in Figure~\ref{fig:skills}. 
%

\subsection{Natural Language Understanding (NLU)}
The NLU module transforms an input text to an input state. 
More concretely, it performs three main tasks: \emph{(i) identifying the domain of an utterance}, \emph{(ii) identifying the intent of an utterance}, and \emph{ (iii) extracting the values of slots that are mentioned in the utterance}. 
\athena\ already features a wide range of NLU models including rule-based and machine learning (ML-)based approaches that not only can be used independently or in combination but also can be easily extended.
%
%
%
\paragraph{Rule-based NLU} 
Our rule-based NLU approach relies on a set of pre-defined templates.  
We overlay templates on a user input text to detect the domain and intent as well as to fill the values of the slots that appear in the matched template. 
A rule-based approach is highly precise in accomplishing NLU tasks, therefore it is suitable for phrases with high frequency, low complexity and low ambiguity. 
However, this approach suffers from low recall.    
If an input text slightly differs from the templates then the NLU module fails to recognize domains and intents.  
Increasing the number of templates does not mitigate this issue but rather fosters ambiguity, for example, the template \emph{When does ... start?} can be used for both \emph{When does \underline{ACL 2020} start?} and \emph{When does \underline{Deep Adversarial Learning for NLP} start?}, where the domain of the former question is \emph{conference} and of the latter one is \emph{tutorial}. 
%
%
%
To overcome this problem, additionally, we train and integrate ML-based NLU models.  
\paragraph{ML-based NLU}
Machine learning models require annotated training data.
Since there is no available dataset for training scientific conversational agents, we propose a new dataset by automatically augmenting our templates with paraphrases.
%
%
%

To do so, we extract 74 most frequent n-grams (n =  \{2,3,4\}) from four non-scientific task-based dialogue corpora: ATIS \cite{hemphill1990atis}, Snips\footnote{snips.ai}, DSTC2 \cite{henderson2014second}, and Frames \cite{asri2017frames}. 
%
%
%
The rationale behind our approach is to capture frequent expressions that are often used by human users to chat with conversational agents (e.g. \emph{give me, I need to know, ...}) and then combine those with the informative parts of our templates (e.g. \emph{the deadline for \{CONF\_NAME\}}) to augment the templates. 
Our augmentation approach makes the ML-based NLU models robust to such variances in the input utterances. 
%
%
The informative expressions are automatically extracted from the templates using two approaches: (i) We extract explicit questions that start with \emph{where, when, which, whose} (e.g., \emph{I don't know when is \{CONF\_NAME\}} $\to$ \emph{when is \{CONF\_NAME\}}), (ii) We extract phrases starting with \emph{the} and \emph{a}, given that the sentence contains \emph{what} or \emph{who} at the beginning\footnote{This ensures that the question is about the slot/event itself and not its location or time or some other attribute} (e.g., \emph{who is the author of \{PAPER\_TITLE\}} $\to$ \emph{the authors of \{PAPER\_TITLE\}}). 
We extract $78$ informative expressions from our templates using the first approach, and $24$ using the second one. 
These phrases can be prepended with extracted most frequent n-grams to produce new templates.   
%
%
%
Finally, for each template, we replace the slots with different concrete slot values obtained from KBs (e.g.,\emph{\{ CONF\_NAME \}} is replaced with different NLP conference names.  
The final number of instances for the training and test sets is shown in Table~\ref{tab:dataset}.
Our dataset is publicly available\footnote{The link comes later}.
\begin{table}[!ht]
\small
\centering
\begin{tabular}{lrl}
                \toprule
                      & \textbf{Train}  &  \textbf{Test}  \\ 
                \midrule
\# of human-provided templates       &   285     &    161     \\
\# of added templates &   1621    &    816    \\ 
\# of instances       &   1906    &    977   \\
                \bottomrule
\end{tabular}
\caption{Some statistics of our dataset for NLU.}
\label{tab:dataset} 
\end{table}
%

%
We use two approaches to encode utterances into vectors. 
First, we represent utterances by their TF-IDF representations as feature-vectors. 
We then supply the vector representation of each utterance to a Support Vector Machine (SVM) with a linear kernel for domain and intent identification, and to a Hidden Markov Model (HMM) for slot filling. 
%
Second, we encode words in a dialogue utterance by GLoVe, as benchmark pre-trained word embeddings, to include the semantic relationships among words. 
We compute the average of word embeddings in an utterance to represent the utterance by a vector. 
%
%
Table~\ref{tab:nlu-results} shows the performance of the described models.  
%
\begin{table}[!ht]
    \small
    \centering
    \begin{tabular}{lll}
        \toprule
             \textbf{Model} &  \textbf{Intent Detection}&  \textbf{Slot Filling} \\ 
        \midrule
            Random baseline   &  $02.67$ & $07.32$ \\
            Majority baseline &  $06.34$ & $64.96$ \\
            HMM  &   -       & $87.20$ \\
            SVM  &   $94.98$  & - \\  
            GolVe-based &   $92.22$  & $98.45$ \\
        \bottomrule
        \end{tabular}
        \caption{The accuracy(\%) of the ML models for NLU.} 
    \label{tab:nlu-results}
\end{table}
Given the results, we use the SVM model for the intent detection and GloVe-based model for slot filling. 
To benefit from both rule-based and ML-based NLU models, we first use the rule-based models to transform an input utterance to a an input state; if it fails then we use the ML-based models.
\subsection{Dialogue Manager (DM)}
%
%
%
We associate each domain with a skill of \athena. 
The dialogue manager is in charge of triggering a sequence of skills to provide a response to an input utterance. 
To do so, we define a tree-based hierarchical relationship among skills (See Figure~\ref{fig:skills}). 
We use the intents in each skill as actions it should perform to react to input utterances.  
Each node in the tree is a skill, which includes several sub-skills to which we refer as ``children''. 
%
%
This sort of hierarchical relation among skills makes developing the DM module efficient because it narrows the scope of the active context at each dialogue turn.
For instance, when a user asks about the title of a paper given its author names, all nodes in the path from the dialogue master, which is the root of the tree, to the \emph{Meta-data} skill are active as a context to provide the response. 
The follow-up question, which could be about showing the abstract of that paper, is interpreted given the active path of the tree as the context. 
However, if the topic of a dialogue changes in the follow-up question, the entire path becomes inactive and another proper path will be activated. 
%
For any input state to DM, it first checks whether there is an active path in the tree. 
If so, it uses the final skill of the active path and gives a response according to the state. 
If not, it will classify the state into a new path.

The tree structure enables \athena\ to consider local context (a few last dialogue turns) for providing a response. 
To benefit from the long history in a conversation, we introduce a memory to retain the most essential information that is given and taken during a conversation with \athena. 
Since the most salient information of input and response utterances are encoded via states, we retain the input and response states of each dialogue turn into a stack of states. 
\subsection{Knowledge Bases (KBs)} 
We design and implement each domain as a skill for \athena\ to ease the process of extending its knowledge for the future use-cases.   
Each skill connects to at least one source of data (which are mainly websites) to acquire relevant data for responding questions.
%
%
Table~\ref{tab:info-source} shows the sources used for extracting data. %
We have one or two sources per domain to demonstrate our approach, while our underlying framework enables implementing connectors to a wide range of additional or alternative sources. 
%
The license terms of these websites give permissions to use their data for research purposes. 
\begin{table}[!ht]
    \resizebox{\columnwidth}{!}{
    \small
    \centering
    \begin{tabular}{@{}ll@{}}
    \toprule
         \textbf{Domain} &  \textbf{Source} \\
    \midrule
        Papers and Authors &  \url{www.aclweb.org/anthology}\\
         &   \url{www.arxiv.org}\\
          & \url{https://scholar.google.com}\\ 
        NLP News &  \url{newsletter.ruder.io}\\
        Conference Deadline &  \url{http://www.wikicfp.com/cfp/}\\
         &  \url{www.aideadlin.es}\\
        Conference Program &  NAACL 2019 database\\
        \bottomrule
    \end{tabular}
    }
    \caption{The data sources used for different domains.}
    \label{tab:info-source}
\end{table}

%
One of the goals of \athena\ is to assist participants in conferences to plan their visit effectively by alleviating the need for searching in the conference programs.  
Such programs present information about the time schedule, location, title, and other details of events (e.g. oral presentation sessions, keynotes, tutorials, etc) in a conference. 
We collect the information related to the conference schedules from their websites. 
%
%
%
%
%
%
%

%
We also implement a script for each of the paper and authors skills to crawl the websites that contain relevant information (See Table\ref{tab:info-source}).
These scripts have two main functions. 
One function obtains the data from a website instantly by making an instance connection to the website and querying the information required for responding to a question.   
This functionality is suitable for KBs that have giant data, e.g., Google Scholar. 
%
The other function downloads essential data from the KBs in a regular time-period.
%
%
%
Besides, to ensure that \athena\ always provides the latest and most real-time information, we let \athena\ update its KBs regularly. 
%
%
%
%

For gathering the deadline dates of NLP and ML conferences we use two source websites (See Table~\ref{tab:info-source}).
Since conference deadlines are set months in advance, we use the capability of our website crawler to collect such data every 30 days from the websites. 
%
%
We design \athena\ to assist researchers in their daily work. 
It is crucial for researcher to know about the latest news in their area. 
Currently, the NLP news website (See Table \ref{tab:info-source}) provides such information. 
\athena\ collects the news from the corresponding website and transforms them to a structured format for further processing, for example summarizing the news. 
%


\subsection{Natural Language Generation (NLG)}
We implement a template-based NLG module that receives the values for its slots from a response state. 
Slot values in response states are obtained from the input state and the information that the corresponding active skill extracts from KBs. 
The task of the NLG module is to find a proper template for generating an informative response.  
To do so, among the templates that are defined for a skill, we filter the most informative ones, which include the most number slots that can be filled. 
We then randomly choose one of these filtered templates to generate a natural response. 
Our motivation of randomly selecting a template is to not repeat a response through a dialogue. 
The filled response templates are first sent to the dialogue master and then transferred to a UI for displaying to users.  
\subsection{Developing New Skills}
We make the framework and the skill set of \athena\ open source with the goal of jointly building a conversational agent for NLP researchers in a community effort. 
Contributing new skills is easy because of our proposed hierarchical structure between skills.  
A skill can easily be added to \athena\ by defining two disjoint sets of NLU and NLG templates.  
%
%
%
We then automatically recognize all new skills, and integrate them into the tree structure. 
\section{Human Evaluation}
To evaluate \athena\ in a realistic scenario, we conduct an in-house user study with one postdoc, three PhD candidates, and two master students from our lab working on different NLP topics. 
Our study consists of three main tasks, i.e.\ intuitiveness, diversity, and information coverage.  
We also collect a general satisfaction survey. 
%
\begin{table*}[!ht]
    \small
    \centering
    \begin{tabular}{@{}p{6cm}cccc@{}}
    \toprule
        & 1 (strongly agree) & 2 (agree) & 3 (disagree) & 4 (strongly disagree) \\
        \midrule
    \athena\ was able to ``understand'' my questions & 16.7\% & \textbf{50.0\%} & 33.3\% & 00.0\% \\ 
     \athena\ was able to provide answers to my questions & 00.0\%  & \textbf{50.0\%} & \textbf{50.0\%} & 00.0\% \\
    I was satisfied with the informativeness of the answers provided by \athena\ & \textbf{33.3\%}  & \textbf{33.3\%} & 00.0\% & \textbf{33.3\%} \\
    I was satisfied with the fluency of the answers provided  by \athena\ &  16.7\% & \textbf{50.0\%} & 16.7\% & 16.7\% \\
    \athena\ could respond in a reasonable time &  \textbf{50.0\%} & 33.3\% & 16.7\% & 00.0\% \\
    The GUI of \athena\ was suitable for reading the provided answers &  \textbf{50.0\%} & 00.0\% & 33.3\% & 16.7\% \\
    \athena\ reduces my need to google a specific information &  16.7\% & \textbf{66.7\%} & 00.0\% & 16.7\% \\
    \athena\ would help me save some time in my work &  \textbf{33.3\%} & \textbf{33.3\%} & \textbf{33.3\%} & 00.0\% \\
    I would like to use \athena\ in the future on a daily basis &  00.0\% & \textbf{66.7\%} & 00.0\% & 33.3\% \\
    I will use \athena\ to plan for my next conference &  16.7\% & 16.7\% & \textbf{33.3\%} & \textbf{33.3\%} \\
    \bottomrule
    \end{tabular}
    \caption{The output of the general survey.}
    \label{tab:heval-survey}
\end{table*}
\paragraph{Intuitiveness} 
We aim at estimating the intuitiveness of \athena\ for the users who interact with it for the first time. 
We provide human judges with a set of slot values as the pieces of information that they can inquiry.
They are asked to randomly choose a piece of information and then keep formulating different questions until \athena\ provides a correct response. 
We measure the average number of formulations users defined to obtain the correct response as a proxy for intuitiveness of \athena. 
Users are asked to stop this task if \athena\ fails to deliver a correct response after 20 tries.  
\paragraph{Diversity}
We measure to what extent \athena\ identifies the intent of various input questions targeting an identical piece of information.  
For this task human judges are provided with a set of information. 
Then they are asked to choose a piece of information (different from the one chosen in the intuitiveness task) to talk to \athena. 
The difference to the intuitiveness task is that human judges are asked to try five different formulations for asking the selected piece of information, regardless of whether \athena\ succeeds or fails in providing the correct response. 
We report the percentage of the formulations for which \athena\ provides correct responses, with respect to the total number of formulations tried by all human judges.  

\paragraph{Information coverage}
We measure how well \athena\ responds questions about different slot values given by human judges.  
We provide a set of information and their corresponding question templates for interacting with \athena. 
However, templates contain one slot, which needs to be replaced with a concrete value of that slot type by human judges. 
To narrow the domain of slot values, we ask human judges to focus on the NLP research area. 
%
We report the percentage of slot values for which \athena\ responded correctly, with respect to the total number of slot values tried by all human judges.
\paragraph{General survey} 
The human judges are asked to participate in a survey (Table~\ref{tab:heval-survey}) after completing the above tasks to assess their general feelings from interacting with \athena. 
They assign integer scores between 1 and 5 to answer the survey questions. 
%
%

\paragraph{Results}
%
%

The evaluation scores for intuitiveness, diversity, and coverage, which are described above, are $3.8$, $45.83\%$, and $37.50\%$, respectively. 
For the intuitiveness task, among 13 given pieces of information, human judges chose \emph{the number of citations of a paper}, \emph{keynote speakers in a conference}, \emph{the conclusions of a paper}, \emph{the deadline of conference (2X)}, and \emph{the abstract of a paper}, where \athena\ responded to the questions about two latter ones correctly in the first try. 
%
For the diversity task, human judges were interested in the following given pieces of information: \emph{the authors of a paper}, \emph{
the conclusions of a paper (2X)}, \emph{the deadline of a conference}, \emph{the tutorials in a conference}, and \emph{the keynote times in conference}. 

For information coverage, the human judges chose the following pieces of information: \emph{the deadline of conference}, \emph{the start time of a keynote at a conference}, \emph{the h-index of a person}, \emph{retrieving the figures (2X) in a paper}, and \emph{the bib entry of a paper}. 
For \emph{h-index}, \athena\ correctly responded all examined slot values (three out of three). 
%

The results of the survey shows a general satisfaction feeling of interactions with \athena, confirming our motivation that the existence of such an agent helps researchers (See Table~\ref{tab:heval-survey}). 
83\% of participants agree that \athena\ reduces their needs to search through the web (e.g.\ using search engines) to obtain information related to their research; and 66\% use \athena\ in the future. 
However, 66\% of human judges disagreed on using \athena\ for planning their schedule for a conference.  
This observation could be because the current version of \athena\ mainly retrieves information for users but planning for a conference needs some inferences on such information as well.  

This observation manifests itself more in the question about the possible future avenues for \athena: What features would be helpful for your daily work and would like to see in \athena? 
We group the answers as follows: (i) \athena\ should gather some background information about its users and their research interest either by asking some questions during conversations or in its login page,  (ii) \athena\ needs to cover more information from web such as social media and the content of papers, and (iii) \athena\ needs to make some inferences on the retrieved information to help users, for example by inferring if a paper is interesting for a user. 

\section{Conclusions}
The size of research communities is drastically growing which yields exploding information about them on the web.
Accessing such an amount of heterogeneous information in a coherent way takes much time and attention of researchers. 
We propose UKP-ATHENA to ease the access to this information through a conversational environment. 
The current version of UKP-ATHENA achieves satisfactory results based on our human evaluations.  
In future work, we would enable UKP-ATHENA to respond questions about the content information of scientific papers and to perform some inference on conference data. 
UKP-ATHENA is publicly available to chat: \url{http://athena.ukp.informatik.tu-darmstadt.de:5002}. 

\bibliography{paperbib}
\bibliographystyle{acl_natbib}

\appendix

\end{document}